\documentclass[utf8]{FrontiersinHarvard} 

\setcitestyle{square} 
\usepackage{url,hyperref,lineno,microtype,subcaption}
\usepackage[onehalfspacing]{setspace}
\usepackage{amsmath,mathtools,amsthm,physics}
\usepackage{algpseudocode,algorithm}
\usepackage{longtable}
\usepackage{graphicx, caption,subcaption}

\def\keyFont{\fontsize{8}{11}\helveticabold }
\def\firstAuthorLast{Pan {et~al.}} 
\def\Authors{Wenhao Pan\,$^{1,*}$, Anil Aswani\,$^{2}$ and Chen Chen\,$^{3}$}
\def\Address{$^{1}$Departments of Statistics and EECS, University of California, Berkeley, Berkeley, California, USA \\
$^{2}$Industrial Engineering and Operations Research, University of California, Berkeley, Berkeley, California, USA \\
$^{3}$Industrial and Systems Engineering, The Ohio State University, Columbus, Ohio, USA}
\def\corrAuthor{Wenhao Pan}

\def\corrEmail{wenhao1102@berkeley.edu}

\newcommand{\RealField}{\mathbb{R}}
\newcommand{\code}{\texttt}

\theoremstyle{definition}
\newtheorem{definition}{Definition}[section]

\theoremstyle{plain}
\newtheorem{proposition}{Proposition}[section]
\newtheorem{corollary}{Corollary}[section]

\DeclareMathOperator{\Rank}{rank}
\DeclareMathOperator{\CvxHull}{conv}

\begin{document}
\onecolumn
\firstpage{1}

\title {Accelerated Nonnegative Tensor Completion via Integer Programming} 

\author[\firstAuthorLast ]{\Authors} 
\address{} 
\correspondance{} 

\extraAuth{}

\maketitle

\begin{abstract}

The problem of tensor completion has applications in healthcare, computer vision, and other domains. However, past approaches to tensor completion have faced a tension in that they either have polynomial-time computation but require exponentially more samples than the information-theoretic rate, or they use fewer samples but require solving NP-hard problems for which there are no known practical algorithms. A recent approach, based on integer programming, resolves this tension for nonnegative tensor completion. It achieves the information-theoretic sample complexity rate and deploys the Blended Conditional Gradients algorithm, which requires a linear (in numerical tolerance) number of oracle steps to converge to the global optimum. The tradeoff in this approach is that, in the worst case, the oracle step requires solving an integer linear program. Despite this theoretical limitation, numerical experiments show that this algorithm can, on certain instances, scale up to 100 million entries while running on a personal computer. The goal of this paper is to further enhance this algorithm, with the intention to expand both the breadth and scale of instances that can be solved. We explore several variants that can maintain the same theoretical guarantees as the algorithm, but offer potentially faster computation. We consider different data structures, acceleration of gradient descent steps, and the use of the Blended Pairwise Conditional Gradients algorithm. We describe the original approach and these variants, and conduct numerical experiments in order to explore various tradeoffs in these algorithmic design choices.

\tiny
 \keyFont{ \section{Keywords:} Tensor Completion, Integer Programming, Conditional Gradient Method, Acceleration, Benchmarking} 
\end{abstract}

\section{Introduction}

A tensor is a multi-dimensional array or a generalized matrix. $\psi$ is called an order-$p$ tensor if $\psi \in \RealField^{r_1 \times \cdots \times r_p}$ where $r_i$ is the length of the $i$-th dimension of $\psi$. For example, an RGB image with a size of $30$ by $30$ pixels is an order-$3$ tensor in $\RealField^{30 \times 30 \times 3}$. Since tensors and matrices are closely related, many matrix problems can be naturally generalized to tensors, such as computing a matrix norm and decomposing a matrix. However, the tensor generalization of such problems can be substantially more computationally challenging \citep{hillar2013most}.

Like matrix completion, tensor completion uses the observed entries of a partially observed tensor $\psi$ to interpolate the missing entries with a restriction on the rank of the interpolated tensor. The purpose of the rank restriction is to restrict the degree of freedom of the missing entries \citep{10.1145/3278607}, e.g. avoiding overfitting. Without this rank restriction, the tensor completion problem becomes ill-posed because there are too many degrees of freedom that need to be constrained by the available data. Tensor completion is a versatile model with many important applications in social sciences \citep{tan2013new}, healthcare \citep{gandy2011tensor}, computer vision \citep{6138863}, and many other domains. 

In the past decade, there have been major advances in matrix completion \citep{zhang2019robust}. However, for general tensor completion there remains a critical tension. Past approaches either have polynomial-time computation but require exponentially more samples than the information-theoretic rate \citep{gandy2011tensor, mu2014square, barak2016noisy, montanari2018spectral}, or they achieve the information-theoretic rate but require solving NP-hard problems for which there are no known practical numerical algorithms \citep{chandrasekaran2012convex, yuan2016tensor, yuan2017incoherent, rauhut2021tensor}. 

We note that, aside from  matrix completion, for some special cases of tensor completion there are numerical algorithms that can achieve the information-theoretic rate; for instance, nonnegative rank-1 \citep{aswani2016low} tensors, or orthogonal symmetric \citep{rao2015forward} tensors. In this paper, we focus on a new approach, proposed by \citep{bugg2022nonnegative}, that is designed for entrywise-nonnegative tensors, which naturally exist in applications such as image demosaicing. The authors defined a new norm for nonnegative tensors by using the gauge of a specific 0-1 polytope that they constructed. By using this gauge norm, their approach achieved the information-theoretic rate in terms of sample complexity, although the resulting problem is NP-hard to solve. Nevertheless, a practical approach was attained: as the norm is defined by using a 0-1 polytope, the authors embedded integer linear optimization within the \textit{Blended Conditional Gradients} (BCG) algorithm \citep{braun2019blended}, a variant of the Frank-Wolfe algorithm, to construct a numerical computation algorithm that required a linear (in numerical tolerance) number of oracle steps to converge to the global optimum.

This paper proposes several acceleration techniques for the original numerical computation algorithm created by \citet{bugg2022nonnegative}; we test multiple techniques in combination to evaluate the configuration that provides the best overall speedup. The motivation is to further improve the implementation on large-scale problems. Our variants can maintain the same theoretical guarantees as the original algorithm, while offering potential speedups. Indeed, our experiments demonstrate that such speedups can be attained consistently across a range of problem instances. Additionally, this paper provides a full description of the original numerical computation algorithm and its coding implementation, details that were omitted in \citep{bugg2022nonnegative}.

We summarize preliminary material and introduce the framework and theory of \citet{bugg2022nonnegative}'s nonnegative tensor completion approach in Section \ref{sec:preli}. Then, we describe the computation algorithm of their approach in Section \ref{sec:algo} and our acceleration techniques in Section \ref{sec:acc-algo}. Numerical experiment results are presented in Section \ref{sec:exp}.

\section{Preliminaries}\label{sec:preli}

Given an order-$p$ tensor $\psi\in\RealField^{r_1\times \cdots \times r_p}$, we refer to its entry with indices $x=(x_1,\dots,x_p)$ as $\psi_x\coloneqq\psi_{x_1,\dots,x_p}$. $x_i\in[r_i]$ is the value of the $i$-th index where $[r_i]\coloneqq\{1,\dots,r_i\}$. We also define $\rho \coloneqq \sum_i r_i$, $\pi \coloneqq \prod_i r_i$, and $\mathcal{R} = [r_1]\times\cdots\times [r_p]$. The \textit{probability simplex} $\Delta^k \coloneqq \CvxHull\{e_1,\dots,e_k\}$ is the convex hull of the coordinate vectors in dimension $k$.

A nonnegative rank-1 tensor $\psi$ is defined as $\psi \coloneqq \theta^{(1)} \otimes \cdots \otimes \theta^{(p)}$ where $\theta^{(k)}\in\RealField_{+}^{r_k}$ are nonnegative vectors. Its entry $\psi_x$ is $\prod_{k=1}^{p}\theta_{x_k}^{(k)}$. \citet{bugg2022nonnegative} defined the ball of nonnegative rank-1 tensors whose maximum entry is $\lambda\in\RealField_{+}$ to be
\begin{equation}
        \mathcal{B}_\lambda = \{\psi\,\colon\,\psi_x = \lambda \cdot \prod_{k=1}^{p}\theta_{x_k}^{(k)},\,\theta_{x_k}^{(k)}\in[0,1],\, \text{for}\, x\in\mathcal{R}\}
\end{equation}
so the nonnegative rank of nonnegative tensor is
\begin{equation}
        \Rank_+(\psi) = \min\{q\,|\,\psi=\sum_{k=1}^{q}\psi^k,\, \psi^k\in\mathcal{B}_\infty\,\text{for}\, k\in[q]\}
\end{equation}
where $\mathcal{B}_\infty = \lim_{\lambda\rightarrow\infty} \mathcal{B}_\lambda$. For a $\lambda\in\RealField_{+}$, consider a finite set of points 
\begin{equation}
    \mathcal{S}_\lambda = \{\psi\,\colon\,\psi_x = \lambda \cdot \prod_{k=1}^{p}\theta_{x_k}^{(k)},\,\theta_{x_k}^{(k)}\in\{0,1\},\, \text{for}\, x\in\mathcal{R}\}
\end{equation}
\citet{bugg2022nonnegative} established the following connection between $\mathcal{B}_\lambda$ and $\mathcal{S}_\lambda$:
\begin{equation}
    \mathcal{C}_\lambda \coloneqq \CvxHull(\mathcal{B}_\lambda) = \CvxHull(\mathcal{S}_\lambda),
\end{equation}
where $\mathcal{C}_\lambda$ is the nonnegative tensor polytope. \citet{bugg2022nonnegative} also presented three implications of this result that are useful to their nonnegative tensor completion approach. First, $\mathcal{C}_\lambda$ is a polytope. Second, the elements of $\mathcal{S}_\lambda$ are the vertices of $\mathcal{C}_\lambda$. Third, the following relationships hold: $\mathcal{B}_\lambda = \lambda\mathcal{B}_1$, $\mathcal{S}_\lambda = \lambda\mathcal{S}_1$, and $\mathcal{C}_\lambda = \lambda\mathcal{C}_1$.

\subsection{Norm for Nonnegative Tensors}
Key to the theoretical guarantee and numerical computation of \citet{bugg2022nonnegative}'s approach is their construction of a new norm for nonnegative tensors using a gauge (or Minkowski functional) construction. 
\begin{definition}[\citet{bugg2022nonnegative}]\label{NonTenNorm}
    The function defined as 
    \begin{equation}
        \norm{\psi}_+ \coloneqq \inf\{\lambda \geq 0\, |\, \psi \in \lambda \mathcal{C}_1\}
    \end{equation}
    is a norm for nonnegative tensors $\psi\in\RealField_{+}^{r_1\times\cdots\times r_p}$.
\end{definition}
This norm has an important property that it can be used as a convex surrogate for tensor rank \citep{bugg2022nonnegative}. In other words, if $\psi$ is a nonnegative tensor, then we have $\norm{\psi}_{\max}\leq\norm{\psi}_+\leq\Rank_+(\psi)\cdot\norm{\psi}_{\max}$. If $\norm{\psi}_+=1$, then $\norm{\psi}_+=\norm{\psi}_{\max}$.

\subsection{Nonnegative Tensor Completion}

For a partially observed order-$p$ tensor $\psi$, let $(x\langle i \rangle, y\langle i \rangle)\in\mathcal{R}\times\RealField$ for $i=1,\dots,n$ denote the indices and value of $n$ observed entries. We assume that an entry can be observed multiple times, so let $U\coloneqq\{x\langle 1 \rangle,\dots,x\langle u \rangle\}$ where $u\leq n$ denote the set of unique indices of observed entries. Since $\norm{\cdot}_+$ is a convex surrogate for tensor rank, we have the following nonnegative tensor completion problem
\begin{equation}\label{NonTenComProb}
    \begin{aligned}
\widehat{\psi}\in\arg\min_{\psi} \quad & \frac{1}{n}\sum_{i=1}^{n}\left(y\langle i \rangle - \psi_{x\langle i \rangle}\right)^2\\
\textrm{s.t.} \quad & \norm{\psi}_+ \leq \lambda\\
    \end{aligned}
\end{equation}
where $\lambda\in\RealField_{+}$ is given and $\widehat{\psi}$ is the completed tensor. The feasible set $\{\psi\,\colon\, \norm{\psi}_+\leq\lambda\}$ is equivalent to $\mathcal{C}_\lambda$ by the norm definition (\ref{NonTenNorm}) and $\mathcal{C}_\lambda = \lambda\mathcal{C}_1$.

Although the problem (\ref{NonTenComProb}) is a convex optimization problem, \citet{bugg2022nonnegative} showed that solving it is NP-hard. Nonetheless, \citet{bugg2022nonnegative} managed to design an efficient numerical computation algorithm of global minima of the problem (\ref{NonTenComProb}) by using its substantial structure of the problem. We will explain their algorithm more after showing their findings on the statistical guarantees of the problem (\ref{NonTenComProb}).

\subsubsection{Statistical Guarantees}

\citet{bugg2022nonnegative} observed that the problem (\ref{NonTenComProb}) is equivalent to a \textit{convex aggregation} \citep{nemirovski2000topics,tsybakov2003optimal,lecue2013empirical} problem for a finite set of functions, so we have the following tight generalization bound for the solution of the problem (\ref{NonTenComProb})
\begin{proposition}[\citep{lecue2013empirical}]
    Suppose $\abs{y}\leq b$ almost surely. Given any $\delta>0$, with probability at least $1-4\delta$ we have that
    \begin{equation}
        \mathbb{E}\bigl((y-\widehat{\psi}_x)^2\bigr) \leq \min_{\varphi\in\mathcal{C}_\lambda}\mathbb{E}\bigl((y-\widehat{\varphi}_x)^2\bigr) + c_0\cdot\max\bigl[b^2,\lambda^2\bigr]\cdot\max\bigl[\zeta_n,\frac{\log(1/\delta)}{n}\bigr],
    \end{equation}
    where $c_0$ is an absolute constant and 
    \begin{equation}\label{zeta_n}
        \zeta_n=\begin{cases}
            \frac{2^\rho}{n}, & \text{if}\, 2^\rho\leq\sqrt{n}\\
            \sqrt{\frac{1}{n}\log\left(\frac{e2^\rho}{\sqrt{n}}\right)}, & \text{if}\, 2^\rho>\sqrt{n}\\
        \end{cases}
    \end{equation}
\end{proposition}

Under specific noise models such as an additive noise model, we have the following corollary to the above proposition combined with the fact that $\norm{\cdot}_+$ is a convex surrogate for tensor rank.
\begin{corollary}[\citep{bugg2022nonnegative}]
    Suppose $\varphi$ is a nonnegative tensor with $\Rank_+(\varphi) = k$ and $\norm{\varphi}_{\max} \leq \mu$. If $(x\langle i \rangle, y\langle i \rangle)$ are independent and identically distributed with $\abs{y\langle i\rangle - \varphi_{x\langle i\rangle}}\leq e$ almost surely and $\mathbb{E}y\langle i \rangle = \varphi_{x\langle i\rangle}$. Then given any $\delta > 0$, with probability at least $1-4\delta$ we have
    \begin{equation}
        \mathbb{E}\bigl((y-\widehat{\psi}_x)^2\bigr) \leq e^2 + c_0\cdot(\mu k + e)^2\cdot \max\bigl[\zeta_n,\frac{\log(1/\delta)}{n}\bigr]
    \end{equation}
    where $\zeta_n$ is as in (\ref{zeta_n}) and $c_0$ is an absolute constant.
\end{corollary}

The two results above show that the problem (\ref{NonTenComProb}) achieves the information-theoretic sample complexity rate when $k = O(1)$. 

\section{Original Computation Algorithm}\label{sec:algo}

Since $\mathcal{C}_1$ is a 0-1 polytope, we can employ integer linear optimization to address the linear separation problem associated with this polytope. Thus, we can apply the Frank-Wolfe algorithm or its variants to solve the problem (\ref{NonTenComProb}) to a desired numerical tolerance. \citet{bugg2022nonnegative} choose the BCG variant for two reasons. 

First, the BCG algorithm can terminate (within numerical tolerance) in a linear number of oracle steps for an optimization problem with a polytope feasible set and a strictly convex objective function over the feasible set. To make the objective function in the problem (\ref{NonTenComProb}) strictly convex, we can reformulate the problem (\ref{NonTenComProb}) by changing its feasible set from $\mathcal{C}_\lambda$ to $\text{Proj}_U(\mathcal{C}_\lambda)$. The implementation of this reformulation is to simply discard the unobserved entries of $\psi$. Second, the weak-separation oracle in the BCG algorithm accommodates early termination of the associated integer linear optimization problem, which is formulated as follows
\begin{equation}\label{IntProgProb}
    \begin{aligned}
\min_{\varphi, \theta} \quad & \langle c, \varphi - \psi\rangle\\
\textrm{s.t.} \quad & \lambda\cdot(1 - p)+\lambda\cdot\sum_{k=1}^{p}\theta_{x_k}^{(k)}\leq\varphi_x& &x\in\mathcal{R}\\
    \quad & 0 \leq \varphi_x\leq\lambda\cdot\theta_{x_k}^{(k)} &k\in[p],\, &x\in\mathcal{R}\\
    \quad & \theta_{x_k}^{(k)}\in\{0,1\} &k\in[p],\, &x\in\mathcal{R}
    \end{aligned}
\end{equation}
The feasible set in the problem above is equivalent to $\mathcal{S}_\lambda$, and the linear constraints above are acquired from standard techniques in integer optimization \citep{hansen1979methods, padberg1989boolean}. \citet{bugg2022nonnegative} also deploy a fast alternating minimization heuristic to solve the weak-separation oracle to avoid (if possible) solving the problem (\ref{IntProgProb}) via integer programming oracle.

Next, we will  fully describe the Python 3 implementation of the BCG algorithm adapted by \citet{bugg2022nonnegative} to solve the problem (\ref{NonTenComProb}) and its major computation components. This detailed account is vital as it will aid in our explanation of how to accelerate this algorithm in the subsequent sections. It also supplements the algorithm description omitted in \citep{bugg2022nonnegative}. Their code is available from \url{https://github.com/WenhaoP/TensorComp}. For brevity, we do not explain the abstract frameworks of the adapted BCG (ABCG) or its major computation components here, but they can be found in \citep{braun2019blended} and \citep{bugg2022nonnegative}.

\subsection{Adapted Blended Conditional Gradients}

The ABCG is implemented as the function \code{nonten} in \code{original\textunderscore nonten.py}. The \textit{inputs} are indices of observed entries (\code{X}), values of observed entries (\code{Y}), dimension (\code{r}), $\lambda$ (\code{lpar}), and numerical tolerance (\code{tol}). The \textit{output} is the completed tensor $\widehat{\psi}$ (\code{sol}). There are three important preparation steps of ABCG. First, it reformulates the problem (\ref{NonTenComProb}) by changing the feasible set from $\mathcal{C}_\lambda$ to $\text{Proj}_U(\mathcal{C}_\lambda)$. Second, it normalizes the feasible set to $\text{Proj}_U(\mathcal{C}_1)$ and scales \code{sol} by $\lambda$ at the end of ABCG. Third, it flattens the iterate (\code{psi\textunderscore q}) and recovers the original dimension of \code{sol} at the end of ABCG. 

We first build the integer programming problem (\ref{IntProgProb}) in Gurobi. The iterate is initialized as a tensor of ones; the active vertex set $\{v_1,\dots,v_k\}$ (\code{Pts} and \code{Vts}) and the active vertex weight set $\{\gamma_1,\dots,\gamma_k\}$ (\code{lamb}) are initialized accordingly. \code{Vts} stores the $\theta^{(i)}$'s for constructing each active vertex in \code{Pts}. $\{v_1,\dots,v_k\}$ are all order-$p$ tensors, but they are vectorized in the code. \code{bestbd} stores the best lower bound for the optimal objective value found so far. It is initialized as $0$, a lower bound on the optimal objective value of problem (\ref{NonTenComProb}). \code{objVal} stores the objective function value of the current iterate. \code{m.\textunderscore gap} stores \textit{half} of the current primal gap estimate, the difference between \code{objVal} and the optimal objective function value \citep{braun2019blended}.

The iteration procedure of BCG is implemented as a while loop. In each iteration, we first compute the gradient of the objective function with respect to the current iterate code{c} (lines 318 - 320). Then, we compute the dot products ($\{\langle \code{lpar*c},v_1\rangle, \dots, \langle \code{lpar*c},v_k\rangle\}$) between the scaled gradient (\code{lpar*c}) and each active vertex, store them in \code{pro}, and use \code{pro} to find the away-step vertex (\code{psi\textunderscore a}) and the Frank-Wolfe-step vertex (\code{psi\textunderscore f}). Based on the test that compares $\langle\code{lpar*c},\code{psi\textunderscore a - psi\textunderscore f}\rangle$ and \code{m.\textunderscore gap}, we either use the simplex gradient descent step (SiGD) or the weak-separation oracle step (LPsep) to find the next iterate. Since we initialize \code{m.\textunderscore gap} as $+\infty$, we always solve the problem (\ref{IntProgProb}) to solve the weak-separation oracle in the first iteration. At the end of each iteration, we update \code{objVal} and \code{bestbd}. The iteration procedure stops when either the primal gap estimate (\code{2*m.\textunderscore gap}) or the largest possible primal gap \code{objVal - bestbd} is smaller than \code{tol}. 

\begin{figure}[h]
    \centering
    \caption{Flowchart of ABCG}
    \label{fig:flowchart}
    \includegraphics[width=\textwidth]{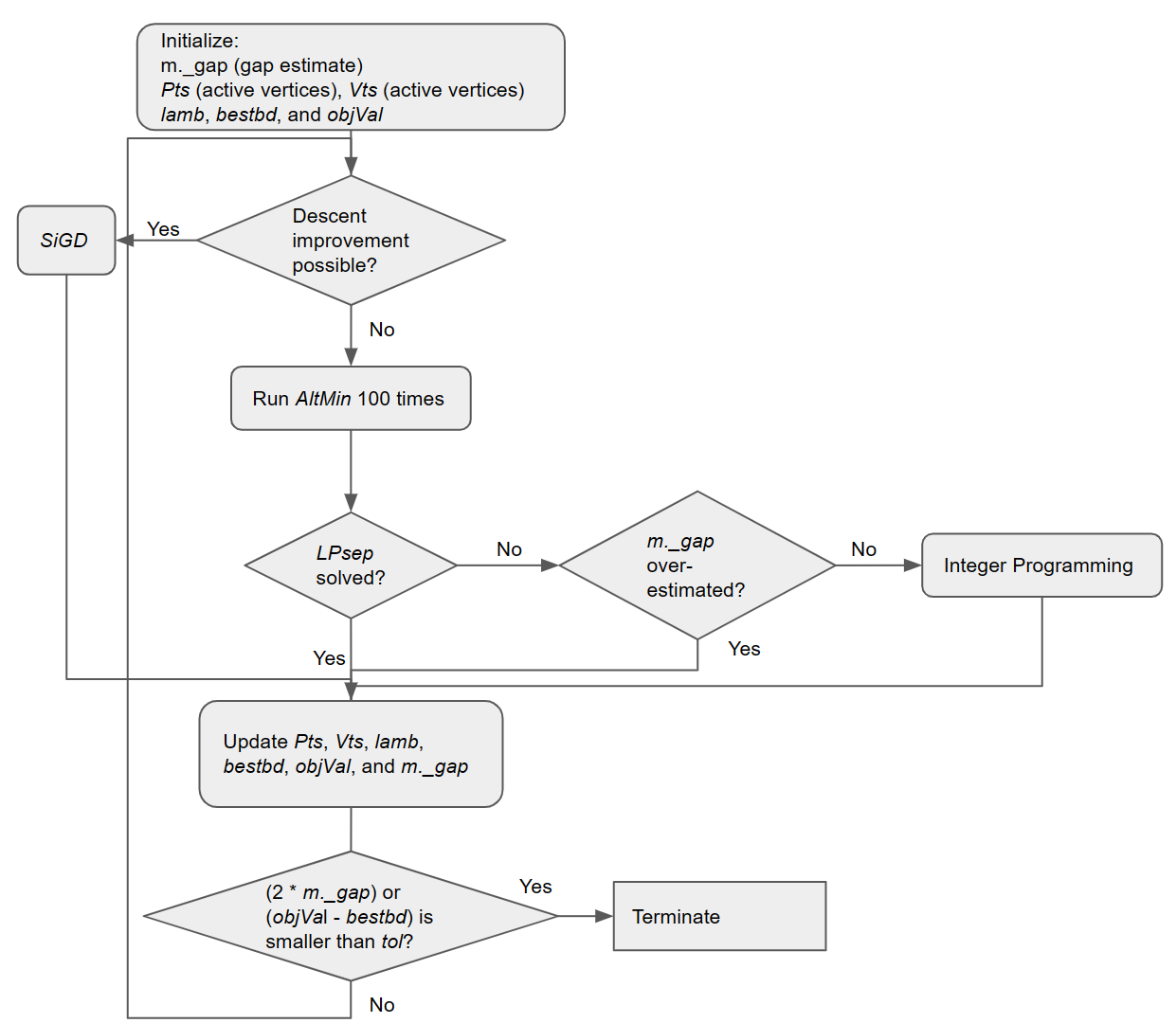}
\end{figure}

Figure \ref{fig:flowchart} is a flowchart that generally explains the entire procedure of the ABCG. 

\subsection{Simplex Gradient Descent Step}

SiGD is implemented in the function \code{nonten}. It determines the next iterate with a lower objective function value via a \textit{single descent step} that solves the following reformulation of the problem (\ref{NonTenComProb})
\begin{equation}
    \begin{aligned}
\hat{\gamma}_1,\dots,\hat{\gamma}_k\in\arg\min_{\gamma_1,\dots,\gamma_k\in\RealField_+} \quad & \frac{1}{n}\sum_{i=1}^{n}\Big(y\langle i \rangle - \big(\sum_{j=1}^{k}\gamma_{j}v_{j}\big)_{x\langle i \rangle}\Big)^2\\
\textrm{s.t.} \quad & \sum_{j=1}^{k}\gamma_j = 1\\
    \end{aligned}
\end{equation}

For clarity, we ignore the projection in the problem above. Since \code{lamb} is in $\Delta^k$ as all its elements are nonnegative and sum to one, SiGD has the term ``Simplex'' in its name \citep{braun2019blended}.

We first compute the projection of \code{pro} onto the hyperplane of $\Delta^k$ and store it in \code{d}. If all elements of \code{d} are zero, then we set the next iterate to the first active vertex and end SiGD. Otherwise, we solve the optimization problem $\hat{\eta} = \arg\max\{\eta\in\RealField_+:\gamma- \eta d\geq0\}$ where $\gamma$ is \code{lamb} and $d$ is \code{d}, and store the optimal solution $\hat{\eta}$ in $\code{eta}$. Next, we compute \code{psi\textunderscore n} which is $x - \hat{\eta}\sum_i d_iv_i$ where $x$ is the current iterate, $d_i$'s are the elements of \code{d}, and $v_i$ are active vertices. If the objective function value of \code{psi\textunderscore n} is smaller or equal to that of the current iterate, then we set the next iterate to \code{psi\textunderscore n} and drop any active vertex with zero weight in the updated active vertex set for constructing \code{psi\textunderscore n}. This is called \textit{drop step} in \citep{braun2019blended}. Otherwise, we perform an exact line search over the line segment between the current iterate and \code{psi\textunderscore n}, and set the next iterate to the optimal solution. This is called \textit{descent step} in \citep{braun2019blended}.

\subsection{Weak-Separation Oracle Step}

LPsep is implemented in the function \code{nonten}. It has two differences from its implementation in the original BCG. First, whether it finds a new vertex that satisfies the weak-separation oracle or not, it always performs an exact line search over the line segment between the current iterate and that new vertex to find the next iterate. Second, it uses \code{bestbd} to update \code{m.\textunderscore gap}.  

\code{m.\textunderscore cmin} is the dot product between \code{lpar*c} and the current iterate. The flag variable \code{oflg} is \code{True} if an improving vertex has not been found that satisfies either the weak-separation oracle or the conditions described below. We first repeatedly use the alternating minimization heuristic (AltMin) to solve the weak-separation oracle. If AltMin fails to do so within 100 repetitions but finds a vertex that shows that \code{m.\textunderscore gap} is an overestimate, we claim that it satisfies the weak-separation oracle and update \code{m.\textunderscore gap}. Otherwise, we solve the integer programming problem (\ref{IntProgProb}) via the Gurobi global solver to obtain a satisfying vertex and update \code{m.\textunderscore gap}. Note that an exact solution is not required; any solution attaining weak separation suffices.

\subsection{Alternating Minimization}

AltMin is implemented as the function \code{altmin} in \code{original\textunderscore nonten.py}. Since a vertex is in $\mathcal{S}_\lambda$, the objective function in the integer programming problem (\ref{IntProgProb}) becomes $\langle c, (\theta^{(1)} \otimes \cdots \otimes \theta^{(p)}) - \psi\rangle$ where $\theta^{(k)}\in\{0,1\}^{r_k}$. There is a motivating observation for solving this problem. For example, if we fix $\theta^{(2)},\dots,\theta^{(p)}$, then the problem (\ref{IntProgProb}) is equivalent to $\min_{\theta^{(1)}\in\{0,1\}^{r_1}} \langle \tilde{c}^{(1)}, \theta^{(1)}\rangle$ where $\tilde{c}$ is a constant in $\RealField^{r_1}$ computed from $c,\theta^{(2)},\dots,\theta^{(p)}$. The optimal solution can be easily found based on the signs of $\tilde{c}$'s entries as $\theta^{(1)}$ is binary. AltMin utilizes this observation.

Given an incumbent solution $\Theta_0\coloneqq\{\hat{\theta}_{0}^{(1)},\dots,\hat{\theta}_{0}^{(p)}\}$, AltMin keeps solving sequences of optimization problems to get $\Theta_1,\Theta_2,\dots,\Theta_T$. For $t=1,\dots,T$ we have
\begin{equation}
    \Theta_t = \{\hat{\theta}_{t}^{(1)} = \arg\min_{\theta_{t}^{(1)}\in\{0,1\}^{r_1}} \langle \tilde{c}_{t}^{(1)}, \theta_{t}^{(1)}\rangle,\, \dots,\, \hat{\theta}_{t}^{(p)}=\arg\min_{\theta_{t}^{(p)}\in\{0,1\}^{r_p}} \langle \tilde{c}_{t}^{(p)}, \theta_{t}^{(p)}\rangle\}.
\end{equation}
where $\tilde{c}_{t}^{(k)}$ stores in the variable \code{fpro}. At each $t$, $\hat{\theta}_{t}^{(1)},\dots,\hat{\theta}_{t}^{(p)}$ are computed one by one. The difference between the objective function values in (\ref{IntProgProb}) of $\Theta_{T-1}$ and $\Theta_T$ are guaranteed to be smaller than \code{tol}. AltMin outputs $\Theta_T$ at the end. 

\section{Accelerated Computation Algorithm} \label{sec:acc-algo}

To show the efficacy and scalability of their approach, \citet{bugg2022nonnegative} conduct numerical experiments on a personal computer. They compare the performance of their approach and three other approaches -- alternating least squares (ALS) \citep{kolda2009tensor}, simple low rank tensor completion (SiLRTC) \citep{6138863}, and trace norm regularized CP decomposition (TNCP) \citep{10.1145/3278607} -- in four sets of different nonnegative tensor completion problems. Normalized mean squared error (NMSE) $\norm*{\widehat{\psi}-\psi}_F^2/\norm*{\psi}_F^2$ is used to measured the accuracy. 

The results of their numerical experiments show that \citet{bugg2022nonnegative}'s approach has higher accuracy but requires more computation time than the three other approaches in all four problem sets. We work directly on their code base to design acceleration techniques. Besides basic acceleration techniques such as caching repetitive computation operations, we design five different acceleration techniques based on profiling results. By combining these five techniques differently, we create ten variants of ABCG that can maintain the same theoretical guarantees but offer potentially faster computation.

\subsection{Technique 1A: Index}

The first technique, \textit{Index}, accelerates AltMin. We initially computed \code{fpro} using a for-loop with $u$ iterations, which can become time-consuming for large $u$ (that is, large sample sizes). To address this, we rewrote the \code{pro} computation to use a for-loop with $r_k$ iterations for $k=1,\dots,p$. Despite each iteration in the new computation being more time-consuming than in the original, our numerical experiments showed a significant decrease in total computation time, particularly for problems with large sample sizes

\subsection{Technique 1B: Pattern}

The second technique, \textit{Pattern}, accelerates AltMin and builds upon the Index technique described above. Instead of extracting information from $U$ during each iteration of the new loop operation in Index, we can extract it at the beginning of the algorithm since $U$ remains unchanged throughout. This approach circumvents unnecessary repetition.

\subsection{Technique 2: Sparse}

The third technique, \textit{Sparse}, accelerates the test that decides whether to use SiGD or LPsep for determining the next iterate. At the beginning of each BCG iteration, \code{Pts} and \code{lpar*c} are used to find \code{psi\textunderscore a} and \code{psi\textunderscore f}. This operation involves a matrix multiplication between \code{Pts} and \code{lpar*c}, which is time-consuming when \code{Pts} has a large size (\textit{i.e.}, large sample size or active vertex set size). We observed the sparsity of \code{Pts} as vertices are binary, and used a SciPy \citep{virtanen2020scipy} sparse matrix to represent it instead of a NumPy \citep{harris2020array} array. It not only accelerates the matrix multiplication but also reduces the memory usage for storing \code{Pts}.

\subsection{Technique 3: NAG}

The fourth technique \textit{NAG} is for accelerating SiGD. Since SiGD is a gradient descent method, we can use Nesterov's accelerated gradient (NAG) to accelerate it. Specifically, we applied \citep{besanccon2022frankwolfe}'s technique of transforming the problem (\ref{NonTenComProb}) to its barycentric coordinates so that we minimize over $\Delta^k$ instead of the convex hull of the active vertex set. We restart the NAG when the active vertex set changes. 

\subsection{Technique 4: BPCG}

The fifth technique \textit{BPCG} is for accelerating SiGD. \citet{tsuji2021sparser} developed the Blended Pairwise Conditional Gradients (BPCG) by combining the Pairwise Conditional Gradients (PCG) with the blending criterion from BCG. Specifically, we applied the lazified version of BPCG, which only differs from BCG by replacing SiGD with PCG.

\subsection{Computation Variants}

We have developed ten computational variants of ABCG by combining the five acceleration techniques described above. Further details about these variants are provided in Table \ref{variants} of Appendix \hyperref[sec:app1]{1}. There are two observations. First, Index and Pattern are always used together as Pattern relies on Index, so we combine and consider them as a single technique in all following discussions. Second, NAG and BPCG are exclusive to each other as BPCG removes SiGD completely. 

\section{Numerical Experiments} \label{sec:exp}

 For benchmarking we adopted the same problems (four sets of nonnegative tensor completion problems) and setup as \citet{bugg2022nonnegative}. We repeated each problem 100 times. For each problem, we constructed the true tensor $\psi$ by randomly choosing 10 points from $\mathcal{S}_1$ and then taking a random convex combination \citep{bugg2022nonnegative}. We conducted the experiments on a laptop equipped with 32GB of RAM and a 2.2Ghz Intel Core i7 processor with 6-cores/12-threads. The algorithms were coded in Python 3. Gurobi v9.5.2 \citep{gurobi} was used to solve the integer programming problem (\ref{IntProgProb}). The code is available from \url{https://github.com/WenhaoP/TensorComp}. We note that the experiments in \citet{bugg2022nonnegative} were conducted using a laptop computer with 8GB of RAM and an Intel Core i5 2.3Ghz processor with 2-cores/4-threads, algorithms coded in Python 3, and Gurobi v9.1 \citep{gurobi} to solve the integer programming problem (\ref{IntProgProb}).

We used NMSE to measure the accuracy and recorded its arithmetic mean (with its standard error) over one hundred repetitions of each problem instance. We recorded the arithmetic mean (with its standard error), geometric mean, minimum, median, and maximum of the computation time over 100 repetitions of each problem instance.  Generally speaking, the most effective variants were version 1 (Index + Pattern) and version 8 (BPCG + Index + Pattern), offering consistent speedups over the original version 0 (ABCG). All variants achieve almost the same NMSE as the original algorithm for each problem, which shows that they maintain in practice the same theoretical guarantee as the original algorithm; moreover, their runtimes were mostly competitive with the original.

\subsection{Experiments with Order-$3$ Tensors}

The first set of problems is order-$3$ tensors with increasing dimensions and $n=500$ samples (with indices sampled with replacement). Table~\ref{order3} in Appendix~\hyperref[sec:app2]{2} reports the timing statistics, including medians. Figure~\ref{fig:first_set} shows arithmetic mean computation time for all variants in the upper panel and for variants 0, 1, and 8 in the lower panel. Among all versions, version 1 (Index + Pattern) has the lowest mean and median computation time for each problem except for the $100 \times 100 \times 100$ tensor problem, where the original version 0 (ABCG) is slightly faster. Thus, we claim version 1 (Index + Pattern) as the \textit{best} to use for this set of problems.  

\subsection{Experiments with Increasing Tensor Order}

\begin{figure}[h!]   
    \centering    
    \caption{Results for increasing order nonnegative tensors with size $10^{\times p}$ and $n = 10,000$ samples.}
    \label{fig:second_set}
    \includegraphics[]{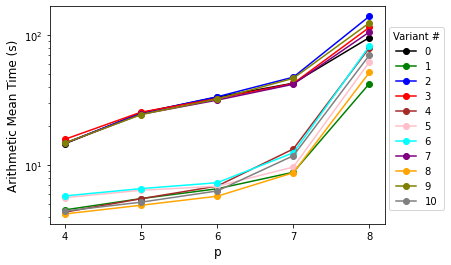}
    \includegraphics[]{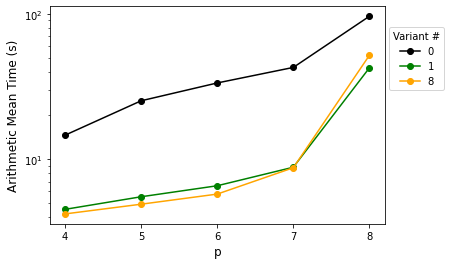}
\end{figure}

The second set of problems is tensors with increasing tensor order, with each dimension $r_i = 10$ for $i=1,\dots,p$, and $n=10,000$ samples. Table~\ref{inc-order} in Appendix~\hyperref[sec:app2]{2} reports the timing statistics, including medians. Figure~\ref{fig:second_set} shows arithmetic mean computation time using the same panel layout as Figure~\ref{fig:first_set}. Among all the versions, version 1 (Index + Pattern) and version 8 (BPCG + Index + Pattern) has the lowest mean and median computation time for each problem except for the $10^{\times 8}$ tensor problem. Thus, we claim versions 1 (Index + Pattern) and 8 (BPCG + Index + Pattern) are the \textit{best} to use for this set of problems.

For the $10^{\times 8}$ tensor problem, version 5 (NAG + Index + Pattern) has the lowest median computation time. We noted that the median computation time for this variant was significantly lower than the mean computation time, a trend also seen with other variants incorporating NAG. The maximum computation time of these versions is also much higher than other versions. This is due to the fact that, in certain cases, AltMin failed to solve the weak-separation oracle, and the Gurobi solver was needed to solve the time-consuming integer programming problem (\ref{IntProgProb}). 

\subsection{Experiments with Increasing Sample Size}

\begin{figure}[h!]
    \centering
    \caption{Results for order-3 nonnegative tensors with size $r \times r \times r$ and $n = 500$ samples.}
    \label{fig:first_set}
    \includegraphics[]{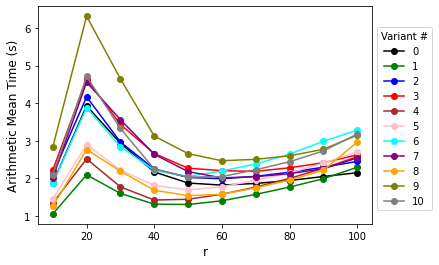}
    \includegraphics[]{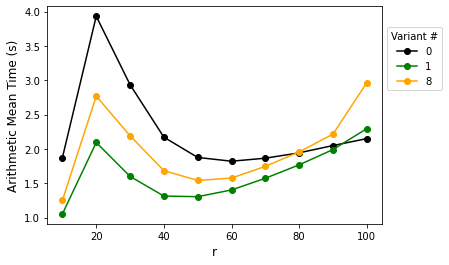}
\end{figure}

\begin{figure}[h!]
    \centering
    \caption{Results for nonnegative tensors with size $10^{\times 6}$ and increasing $n$ samples.}
    \label{fig:third_set}
    \includegraphics[]{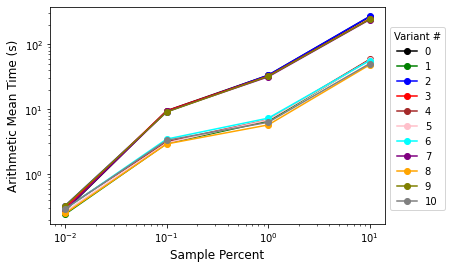}
    \includegraphics[]{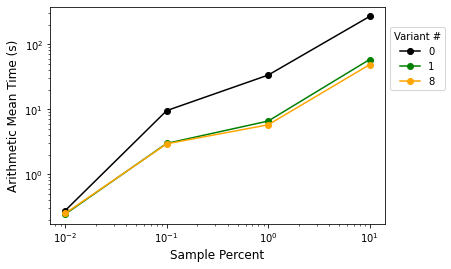}
\end{figure}

\begin{figure}[h!] 
    \centering
    \caption{Results for nonnegative tensors with size $10^{\times 7}$ and increasing $n$ samples.}
    \label{fig:fourth_set}
    \includegraphics[]{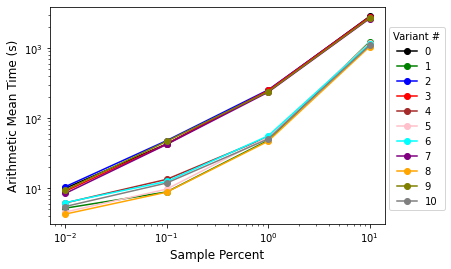}
    \includegraphics[]{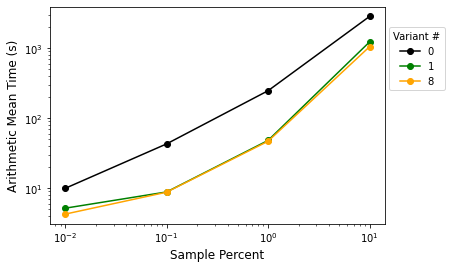}
\end{figure}

The third and fourth set of problems is tensors of size $10^{\times 6}$ and $10^{\times 7}$ with increasing sample sizes. Tables~\ref{inc-n-1} and~\ref{inc-n-2} in Appendix~\hyperref[sec:app2]{2} report the timing statistics, including medians. Figures~\ref{fig:third_set} and~\ref{fig:fourth_set} show arithmetic mean computation time for the two tensor sizes, respectively, using the same panel layout as Figure~\ref{fig:first_set}.  Among all the versions, either version 1 (Index + Pattern) or version 8 (BPCG + Index + Pattern) has the lowest mean and median computation time for each problem. Thus, we claim versions 1 (Index + Pattern) and 8 (BPCG + Index + Pattern) are the \textit{best} to use for this set of problems.

\clearpage

\clearpage

\clearpage

\clearpage

\section{Discussion and Conclusion} \label{sec:con}

 In this paper, we proposed and evaluated a range of speedup techniques for the original numerical computation algorithm for nonnegative tensor completion, which was initially developed by \citet{bugg2022nonnegative}. We benchmarked these algorithm variants on the same set of problem instances designed by \citet{bugg2022nonnegative}. Our benchmarking results were that versions 1 (Index + Pattern) and 8 (BPCG + Index + Pattern) generally had the fastest computation time for solving the nonnegative tensor completion problem (\ref{NonTenComProb}), offering substantial speedups over the original algorithm. Version 1 had Index and Pattern techniques, and version 8 had BPCG, Index, and Pattern techniques. Because version 8 is version 1 with BPCG, version 8 may  be preferred over version 1 for problem instances where the active vertex set could be large. Surprisingly, we found that Sparse failed to yield improvements. A possible reason is that the implementation of Sparse needs extra operations (\textit{e.g.}, finding the indices of nonzero entries) for converting NumPy arrays to SciPy sparse arrays. If the active vertex set is not large enough so that extra operations take time as least as the time saved from matrix multiplications of the active vertex set, then we cannot see an improvement. These results suggest that Index and Pattern are the most important for acceleration, and that BCG and BPCG work equally well overall.

\section*{Conflict of Interest Statement}

The authors declare that the research was conducted in the absence of any commercial or financial relationships that could be construed as a potential conflict of interest.

\section*{Author Contributions}

WP, AA, and CC contributed to the conception and design of the study. WP did the programming, ran the numerical experiments, analyzed the results, and wrote the first draft of the manuscript. WP, AA, and CC contributed to the manuscript revision, read, and approved the submitted version.

\section*{Funding}

This material is based upon work partially supported by the NSF under grant CMMI-1847666.



\section*{Data Availability Statement}
The datasets generated for this study can be found on GitHub: \url{https://github.com/WenhaoP/TensorComp}.

\bibliographystyle{Frontiers-Harvard} 
\bibliography{frontiers}


\newpage
\section*{Appendix 1: Table of Variants}\label{sec:app1} 
\begin{center}
\begin{longtable}{c|c|c|c|c|c}
    \caption{ABCG and its ten variants}\label{variants}\\
    \noalign{\vskip 1mm} \hline \noalign{\vskip 1mm}     
    \textbf{Version} & \textbf{Sparse} & \textbf{NAG} & \textbf{BPCG} & \textbf{Index} & \textbf{Pattern} \\ \noalign{\vskip 1mm} \hline \noalign{\vskip 1mm}     
    0 (ABCG) & FALSE  & FALSE & FALSE & FALSE & FALSE    \\
    1              & FALSE  & FALSE & FALSE & TRUE  & TRUE    \\
    2              & TRUE   & FALSE & FALSE & FALSE & FALSE   \\
    3              & FALSE  & TRUE  & FALSE & FALSE & FALSE   \\
    4              & TRUE   & FALSE & FALSE & TRUE  & TRUE    \\
    5              & FALSE  & TRUE  & FALSE & TRUE  & TRUE    \\
    6              & TRUE   & TRUE  & FALSE & TRUE  & TRUE    \\
    7              & FALSE  & FALSE & TRUE  & FALSE & FALSE   \\
    8              & FALSE  & FALSE & TRUE  & TRUE  & TRUE    \\
    9              & TRUE   & FALSE & TRUE  & FALSE & FALSE   \\
    10             & TRUE   & FALSE & TRUE  & TRUE  & TRUE    \\ \noalign{\vskip 1mm} \hline \noalign{\vskip 1mm}   
    \multicolumn{6}{p{10cm}}{Note: \code{True} means we implement this technique in this variant, and \code{False} means we do not implement this technique in this variant. For example, version 5 implements NAG, Index, and Pattern.}
\end{longtable}  
\end{center}

\newpage
\section*{Appendix 2: Tables of Numerical Experiment Results}\label{sec:app2}
\begin{center}
\begin{longtable}{c|c|c|ccccc}
    \caption{Results for order-$3$ nonnegative tensors with $n = 500$ samples}\label{order3} \\
    \noalign{\vskip 1mm} \hline \noalign{\vskip 1mm}     
    \textbf{Tensor} & \textbf{Version} & \textbf{NMSE} & \multicolumn{5}{c}{\textbf{Time (s)}} \\ 
    \textbf{Size} & ~ & Arithmetic & Arithmetic & Geometric  &  &  & \\
    ~ & ~ & Mean $\pm$ SE & Mean $\pm$ SE &  Mean & Min & Median & Max \\ \noalign{\vskip 1mm} \hline \noalign{\vskip 1mm}    
    $10^{\times 3}$ & 0 & 0.017 $\pm$ 0.001 & 1.874 $\pm$ 0.061 & 1.777 & 0.756 & 1.849 & 3.935 \\ 
        ~ & 1 & 0.017 $\pm$ 0.001 & \textbf{1.054} $\pm$ 0.044 & \textbf{0.966} & 0.352 & \textbf{1.006} & 2.688 \\ 
        ~ & 2 & 0.018 $\pm$ 0.001 & 2.116 $\pm$ 0.071 & 1.996 & 0.757 & 2.102 & 4.114 \\ 
        ~ & 3 & 0.017 $\pm$ 0.001 & 2.225 $\pm$ 0.067 & 2.120 & 0.885 & 2.224 & 4.234 \\ 
        ~ & 4 & 0.018 $\pm$ 0.001 & 1.346 $\pm$ 0.060 & 1.218 & 0.381 & 1.277 & 2.998 \\ 
        ~ & 5 & 0.017 $\pm$ 0.001 & 1.450 $\pm$ 0.056 & 1.339 & 0.398 & 1.386 & 3.168 \\ 
        ~ & 6 & 0.017 $\pm$ 0.001 & 1.868 $\pm$ 0.073 & 1.720 & 0.520 & 1.810 & 4.035 \\ 
        ~ & 7 & 0.017 $\pm$ 0.001 & 2.014 $\pm$ 0.063 & 1.914 & 0.819 & 2.001 & 3.383 \\ 
        ~ & 8 & 0.017 $\pm$ 0.001 & 1.252 $\pm$ 0.054 & 1.135 & 0.443 & 1.211 & 2.956 \\ 
        ~ & 9 & 0.017 $\pm$ 0.001 & 2.834 $\pm$ 0.113 & 2.600 & 0.829 & 2.681 & 5.694 \\ 
        ~ & 10 & 0.017 $\pm$ 0.001 & 2.093 $\pm$ 0.103 & 1.828 & 0.480 & 1.821 & 4.795 \\ \noalign{\vskip 1mm} \hline \noalign{\vskip 1mm}
        $20^{\times 3}$ & 0 & 0.115 $\pm$ 0.004 & 3.931 $\pm$ 0.073 & 3.860 & 2.086 & 3.952 & 6.390 \\ 
        ~ & 1 & 0.115 $\pm$ 0.004 & \textbf{2.094} $\pm$ 0.043 & \textbf{2.046} & 1.020 & \textbf{2.139} & 3.314 \\ 
        ~ & 2 & 0.118 $\pm$ 0.004 & 4.170 $\pm$ 0.078 & 4.091 & 1.884 & 4.272 & 5.912 \\ 
        ~ & 3 & 0.118 $\pm$ 0.004 & 4.639 $\pm$ 0.077 & 4.574 & 2.955 & 4.728 & 7.014 \\ 
        ~ & 4 & 0.118 $\pm$ 0.004 & 2.518 $\pm$ 0.051 & 2.459 & 1.029 & 2.588 & 3.625 \\ 
        ~ & 5 & 0.118 $\pm$ 0.004 & 2.898 $\pm$ 0.051 & 2.849 & 1.522 & 2.980 & 3.958 \\ 
        ~ & 6 & 0.118 $\pm$ 0.004 & 3.879 $\pm$ 0.070 & 3.811 & 1.999 & 3.915 & 5.515 \\ 
        ~ & 7 & 0.114 $\pm$ 0.004 & 4.561 $\pm$ 0.076 & 4.497 & 2.405 & 4.636 & 6.781 \\ 
        ~ & 8 & 0.114 $\pm$ 0.004 & 2.768 $\pm$ 0.049 & 2.722 & 1.173 & 2.742 & 4.529 \\ 
        ~ & 9 & 0.117 $\pm$ 0.004 & 6.315 $\pm$ 0.116 & 6.200 & 3.526 & 6.309 & 9.101 \\ 
        ~ & 10 & 0.117 $\pm$ 0.004 & 4.712 $\pm$ 0.096 & 4.602 & 2.274 & 4.747 & 6.609 \\ \noalign{\vskip 1mm} \hline \noalign{\vskip 1mm}
        $30^{\times 3}$ & 0 & 0.240 $\pm$ 0.005 & 2.935 $\pm$ 0.068 & 2.852 & 1.547 & 2.952 & 4.793 \\ 
        ~ & 1 & 0.240 $\pm$ 0.005 & \textbf{1.600} $\pm$ 0.039 & \textbf{1.551} & 0.865 & \textbf{1.584} & 2.488 \\ 
        ~ & 2 & 0.241 $\pm$ 0.005 & 2.967 $\pm$ 0.077 & 2.871 & 1.659 & 2.931 & 5.475 \\ 
        ~ & 3 & 0.241 $\pm$ 0.005 & 3.435 $\pm$ 0.072 & 3.360 & 1.818 & 3.424 & 5.760 \\ 
        ~ & 4 & 0.241 $\pm$ 0.005 & 1.774 $\pm$ 0.051 & 1.705 & 0.973 & 1.711 & 3.659 \\ 
        ~ & 5 & 0.241 $\pm$ 0.005 & 2.222 $\pm$ 0.053 & 2.161 & 1.228 & 2.160 & 3.936 \\ 
        ~ & 6 & 0.241 $\pm$ 0.005 & 2.846 $\pm$ 0.073 & 2.757 & 1.600 & 2.757 & 5.172 \\ 
        ~ & 7 & 0.240 $\pm$ 0.005 & 3.561 $\pm$ 0.082 & 3.467 & 1.808 & 3.419 & 5.874 \\ 
        ~ & 8 & 0.240 $\pm$ 0.005 & 2.192 $\pm$ 0.058 & 2.112 & 1.060 & 2.142 & 3.501 \\ 
        ~ & 9 & 0.239 $\pm$ 0.005 & 4.649 $\pm$ 0.139 & 4.433 & 2.079 & 4.598 & 8.300 \\ 
        ~ & 10 & 0.239 $\pm$ 0.005 & 3.328 $\pm$ 0.117 & 3.112 & 1.475 & 3.313 & 5.927 \\ \noalign{\vskip 1mm} \hline \noalign{\vskip 1mm}
        $40^{\times 3}$ & 0 & 0.323 $\pm$ 0.004 & 2.171 $\pm$ 0.053 & 2.111 & 1.377 & 2.099 & 4.007 \\ 
        ~ & 1 & 0.323 $\pm$ 0.004 & \textbf{1.314} $\pm$ 0.032 & \textbf{1.279} & 0.865 & \textbf{1.255} & 2.528 \\ 
        ~ & 2 & 0.322 $\pm$ 0.004 & 2.250 $\pm$ 0.053 & 2.193 & 1.439 & 2.214 & 4.380 \\ 
        ~ & 3 & 0.321 $\pm$ 0.004 & 2.661 $\pm$ 0.053 & 2.610 & 1.723 & 2.676 & 4.692 \\ 
        ~ & 4 & 0.322 $\pm$ 0.004 & 1.423 $\pm$ 0.033 & 1.389 & 0.926 & 1.365 & 2.847 \\ 
        ~ & 5 & 0.321 $\pm$ 0.004 & 1.812 $\pm$ 0.036 & 1.780 & 1.225 & 1.755 & 3.351 \\ 
        ~ & 6 & 0.321 $\pm$ 0.004 & 2.211 $\pm$ 0.045 & 2.172 & 1.546 & 2.114 & 4.350 \\ 
        ~ & 7 & 0.321 $\pm$ 0.004 & 2.641 $\pm$ 0.059 & 2.577 & 1.484 & 2.627 & 4.666 \\ 
        ~ & 8 & 0.321 $\pm$ 0.004 & 1.684 $\pm$ 0.042 & 1.636 & 0.984 & 1.626 & 3.312 \\ 
        ~ & 9 & 0.321 $\pm$ 0.004 & 3.125 $\pm$ 0.083 & 3.022 & 1.911 & 3.034 & 5.673 \\ 
        ~ & 10 & 0.321 $\pm$ 0.004 & 2.248 $\pm$ 0.071 & 2.151 & 1.333 & 2.058 & 4.529 \\ \noalign{\vskip 1mm} \hline \noalign{\vskip 1mm}
        $50^{\times 3}$ & 0 & 0.385 $\pm$ 0.005 & 1.876 $\pm$ 0.042 & 1.835 & 1.327 & 1.725 & 3.046 \\ 
        ~ & 1 & 0.385 $\pm$ 0.005 & \textbf{1.305} $\pm$ 0.027 & \textbf{1.281} & 0.900 & \textbf{1.216} & 2.289 \\ 
        ~ & 2 & 0.383 $\pm$ 0.004 & 2.025 $\pm$ 0.045 & 1.982 & 1.321 & 1.885 & 3.622 \\ 
        ~ & 3 & 0.382 $\pm$ 0.005 & 2.272 $\pm$ 0.047 & 2.229 & 1.676 & 2.162 & 3.866 \\ 
        ~ & 4 & 0.383 $\pm$ 0.004 & 1.444 $\pm$ 0.029 & 1.418 & 1.077 & 1.360 & 2.535 \\ 
        ~ & 5 & 0.382 $\pm$ 0.005 & 1.691 $\pm$ 0.032 & 1.663 & 1.248 & 1.618 & 2.760 \\ 
        ~ & 6 & 0.382 $\pm$ 0.005 & 2.053 $\pm$ 0.035 & 2.026 & 1.566 & 1.974 & 3.193 \\ 
        ~ & 7 & 0.384 $\pm$ 0.004 & 2.181 $\pm$ 0.054 & 2.121 & 1.447 & 2.010 & 3.839 \\ 
        ~ & 8 & 0.384 $\pm$ 0.004 & 1.541 $\pm$ 0.038 & 1.499 & 1.055 & 1.425 & 2.697 \\ 
        ~ & 9 & 0.384 $\pm$ 0.005 & 2.659 $\pm$ 0.060 & 2.601 & 1.889 & 2.459 & 4.401 \\ 
        ~ & 10 & 0.384 $\pm$ 0.005 & 2.033 $\pm$ 0.046 & 1.989 & 1.504 & 1.897 & 3.544 \\ \noalign{\vskip 1mm} \hline \noalign{\vskip 1mm}
        $60^{\times 3}$ & 0 & 0.444 $\pm$ 0.004 & 1.821 $\pm$ 0.031 & 1.797 & 1.409 & 1.733 & 2.765 \\ 
        ~ & 1 & 0.444 $\pm$ 0.004 & \textbf{1.403} $\pm$ 0.023 & \textbf{1.386} & 1.079 & \textbf{1.342} & 2.154 \\ 
        ~ & 2 & 0.444 $\pm$ 0.004 & 1.990 $\pm$ 0.033 & 1.965 & 1.550 & 1.887 & 2.965 \\ 
        ~ & 3 & 0.444 $\pm$ 0.004 & 2.205 $\pm$ 0.044 & 2.166 & 1.565 & 2.013 & 3.808 \\ 
        ~ & 4 & 0.444 $\pm$ 0.004 & 1.577 $\pm$ 0.024 & 1.560 & 1.213 & 1.504 & 2.348 \\ 
        ~ & 5 & 0.444 $\pm$ 0.004 & 1.777 $\pm$ 0.031 & 1.751 & 1.309 & 1.641 & 2.991 \\ 
        ~ & 6 & 0.444 $\pm$ 0.004 & 2.197 $\pm$ 0.035 & 2.171 & 1.626 & 2.078 & 3.722 \\ 
        ~ & 7 & 0.445 $\pm$ 0.005 & 2.013 $\pm$ 0.044 & 1.970 & 1.442 & 1.879 & 3.631 \\ 
        ~ & 8 & 0.445 $\pm$ 0.005 & 1.576 $\pm$ 0.032 & 1.546 & 1.138 & 1.478 & 2.842 \\ 
        ~ & 9 & 0.446 $\pm$ 0.004 & 2.469 $\pm$ 0.049 & 2.427 & 1.791 & 2.322 & 4.521 \\ 
        ~ & 10 & 0.446 $\pm$ 0.004 & 2.047 $\pm$ 0.041 & 2.012 & 1.483 & 1.953 & 3.967 \\ \noalign{\vskip 1mm} \hline \noalign{\vskip 1mm}
        $70^{\times 3}$ & 0 & 0.497 $\pm$ 0.004 & 1.865 $\pm$ 0.032 & 1.841 & 1.386 & 1.791 & 3.426 \\ 
        ~ & 1 & 0.497 $\pm$ 0.004 & \textbf{1.573} $\pm$ 0.027 & \textbf{1.552} & 1.174 & \textbf{1.520} & 2.861 \\ 
        ~ & 2 & 0.498 $\pm$ 0.004 & 2.054 $\pm$ 0.032 & 2.031 & 1.577 & 1.995 & 3.699 \\ 
        ~ & 3 & 0.495 $\pm$ 0.005 & 2.187 $\pm$ 0.042 & 2.152 & 1.548 & 2.084 & 4.448 \\ 
        ~ & 4 & 0.498 $\pm$ 0.004 & 1.768 $\pm$ 0.026 & 1.751 & 1.342 & 1.728 & 3.095 \\ 
        ~ & 5 & 0.495 $\pm$ 0.005 & 1.947 $\pm$ 0.036 & 1.920 & 1.332 & 1.895 & 4.123 \\ 
        ~ & 6 & 0.495 $\pm$ 0.005 & 2.384 $\pm$ 0.039 & 2.357 & 1.689 & 2.332 & 4.698 \\ 
        ~ & 7 & 0.498 $\pm$ 0.004 & 2.045 $\pm$ 0.043 & 2.006 & 1.480 & 1.933 & 3.964 \\ 
        ~ & 8 & 0.498 $\pm$ 0.004 & 1.746 $\pm$ 0.037 & 1.713 & 1.246 & 1.640 & 3.412 \\ 
        ~ & 9 & 0.500 $\pm$ 0.004 & 2.507 $\pm$ 0.055 & 2.462 & 1.891 & 2.387 & 5.654 \\ 
        ~ & 10 & 0.500 $\pm$ 0.004 & 2.229 $\pm$ 0.049 & 2.190 & 1.694 & 2.118 & 5.110 \\ \noalign{\vskip 1mm} \hline \noalign{\vskip 1mm}
        $80^{\times 3}$ & 0 & 0.546 $\pm$ 0.004 & 1.943 $\pm$ 0.036 & 1.915 & 1.439 & 1.856 & 3.640 \\ 
        ~ & 1 & 0.546 $\pm$ 0.004 & \textbf{1.768} $\pm$ 0.031 & \textbf{1.743} & 1.32 & \textbf{1.700} & 3.273 \\ 
        ~ & 2 & 0.545 $\pm$ 0.004 & 2.163 $\pm$ 0.035 & 2.138 & 1.665 & 2.092 & 3.992 \\ 
        ~ & 3 & 0.542 $\pm$ 0.004 & 2.282 $\pm$ 0.041 & 2.249 & 1.654 & 2.158 & 3.858 \\ 
        ~ & 4 & 0.545 $\pm$ 0.004 & 1.998 $\pm$ 0.032 & 1.976 & 1.552 & 1.935 & 3.759 \\ 
        ~ & 5 & 0.542 $\pm$ 0.004 & 2.140 $\pm$ 0.037 & 2.112 & 1.566 & 2.043 & 3.686 \\ 
        ~ & 6 & 0.542 $\pm$ 0.004 & 2.645 $\pm$ 0.041 & 2.617 & 1.959 & 2.568 & 4.670 \\ 
        ~ & 7 & 0.547 $\pm$ 0.004 & 2.125 $\pm$ 0.044 & 2.087 & 1.589 & 2.002 & 4.312 \\ 
        ~ & 8 & 0.547 $\pm$ 0.004 & 1.956 $\pm$ 0.043 & 1.920 & 1.430 & 1.859 & 4.544 \\ 
        ~ & 9 & 0.546 $\pm$ 0.004 & 2.600 $\pm$ 0.054 & 2.557 & 1.902 & 2.484 & 6.003 \\ 
        ~ & 10 & 0.546 $\pm$ 0.004 & 2.443 $\pm$ 0.050 & 2.405 & 1.858 & 2.371 & 5.699 \\ \noalign{\vskip 1mm} \hline \noalign{\vskip 1mm}
        $90^{\times 3}$ & 0 & 0.580 $\pm$  0.004 & 2.049 $\pm$ 0.039 & 2.016 & 1.416 & 1.928 & 3.341 \\ 
        ~ & 1 & 0.580 $\pm$  0.004 & \textbf{1.991} $\pm$ 0.038 & \textbf{1.959} & 1.441 & \textbf{1.876} & 3.380 \\ 
        ~ & 2 & 0.580 $\pm$  0.004 & 2.296 $\pm$ 0.036 & 2.271 & 1.663 & 2.216 & 3.516 \\ 
        ~ & 3 & 0.575 $\pm$  0.004 & 2.416 $\pm$ 0.037 & 2.387 & 1.669 & 2.366 & 3.210 \\ 
        ~ & 4 & 0.580 $\pm$  0.004 & 2.269 $\pm$ 0.035 & 2.244 & 1.697 & 2.189 & 3.367 \\ 
        ~ & 5 & 0.575 $\pm$  0.004 & 2.404 $\pm$ 0.038 & 2.375 & 1.671 & 2.367 & 3.488 \\ 
        ~ & 6 & 0.575 $\pm$  0.004 & 2.987 $\pm$ 0.042 & 2.958 & 2.101 & 2.987 & 4.231 \\ 
        ~ & 7 & 0.579 $\pm$  0.004 & 2.257 $\pm$ 0.045 & 2.216 & 1.541 & 2.108 & 3.658 \\ 
        ~ & 8 & 0.579 $\pm$  0.004 & 2.215 $\pm$ 0.045 & 2.174 & 1.474 & 2.062 & 3.642 \\ 
        ~ & 9 & 0.579 $\pm$  0.004 & 2.774 $\pm$ 0.042 & 2.744 & 1.854 & 2.693 & 4.205 \\ 
        ~ & 10 & 0.579 $\pm$  0.004 & 2.729 $\pm$ 0.043 & 2.698 & 1.886 & 2.679 & 4.089 \\ \noalign{\vskip 1mm} \hline \noalign{\vskip 1mm}
        $100^{\times 3}$ & 0 & 0.611 $\pm$ 0.004 & \textbf{2.150} $\pm$ 0.038 & \textbf{2.119} & 1.520 & \textbf{2.074} & 3.353 \\ 
        ~ & 1 & 0.611 $\pm$ 0.004 & 2.295 $\pm$ 0.041 & 2.262 & 1.660 & 2.195 & 3.702 \\ 
        ~ & 2 & 0.611 $\pm$ 0.004 & 2.451 $\pm$ 0.037 & 2.425 & 1.801 & 2.373 & 3.946 \\ 
        ~ & 3 & 0.603 $\pm$ 0.004 & 2.624 $\pm$ 0.054 & 2.575 & 1.671 & 2.484 & 4.746 \\ 
        ~ & 4 & 0.611 $\pm$ 0.004 & 2.575 $\pm$ 0.039 & 2.547 & 1.839 & 2.530 & 4.168 \\ 
        ~ & 5 & 0.603 $\pm$ 0.004 & 2.708 $\pm$ 0.054 & 2.659 & 1.815 & 2.589 & 4.811 \\ 
        ~ & 6 & 0.603 $\pm$ 0.004 & 3.285 $\pm$ 0.059 & 3.237 & 2.173 & 3.170 & 5.938 \\ 
        ~ & 7 & 0.608 $\pm$ 0.004 & 2.542 $\pm$ 0.065 & 2.468 & 1.645 & 2.380 & 4.636 \\ 
        ~ & 8 & 0.608 $\pm$ 0.004 & 2.963 $\pm$ 0.097 & 2.845 & 1.831 & 2.698 & 8.809 \\ 
        ~ & 9 & 0.609 $\pm$ 0.004 & 3.149 $\pm$ 0.061 & 3.097 & 2.128 & 3.005 & 6.003 \\ 
        ~ & 10 & 0.609 $\pm$ 0.004 & 3.178 $\pm$ 0.062 & 3.126 & 2.027 & 3.029 & 6.136 \\  \noalign{\vskip 1mm} \hline \noalign{\vskip 1mm}  
        \multicolumn{8}{p{15cm}}{Note: For each problem, the lowest mean and median computation time are highlighted.}  
\end{longtable}
\end{center}

\newpage

\begin{center}
\begin{longtable}{c|c|c|ccccc}
    \caption{Results for increasing order nonnegative tensors and $n = 10,000$ samples}\label{inc-order} \\
    \noalign{\vskip 1mm} \hline \noalign{\vskip 1mm}     
     \textbf{Tensor} & \textbf{Version} & \textbf{NMSE} & \multicolumn{5}{c}{\textbf{Time (s)}} \\ 
    \textbf{Size} & ~ & Arithmetic & Arithmetic & Geometric  &  &  & \\
    ~ & ~ & Mean $\pm$ SE & Mean $\pm$ SE &  Mean & Min & Median & Max \\ \noalign{\vskip 1mm} \hline \noalign{\vskip 1mm}    
    $10^{\times 4}$ & 0 & 0.010 $\pm$ 0.000 & 14.620 $\pm$ 0.160 & 14.533 & 11.13 & 14.723 & 20.162 \\ 
        ~ & 1 & 0.010 $\pm$ 0.000 & 4.515 $\pm$ 0.081 & 4.443 & 2.658 & 4.497 & 7.198 \\ 
        ~ & 2 & 0.010 $\pm$ 0.000 & 14.736 $\pm$ 0.194 & 14.622 & 11.543 & 14.569 & 25.271 \\ 
        ~ & 3 & 0.010 $\pm$ 0.000 & 15.814 $\pm$ 0.194 & 15.698 & 11.103 & 15.732 & 21.805 \\ 
        ~ & 4 & 0.010 $\pm$ 0.000 & 4.335 $\pm$ 0.084 & 4.260 & 2.725 & 4.237 & 8.669 \\ 
        ~ & 5 & 0.010 $\pm$ 0.000 & 5.608 $\pm$ 0.098 & 5.521 & 2.963 & 5.590 & 8.890 \\ 
        ~ & 6 & 0.010 $\pm$ 0.000 & 5.784 $\pm$ 0.107 & 5.686 & 2.974 & 5.752 & 9.049 \\ 
        ~ & 7 & 0.010 $\pm$ 0.000 & 14.761 $\pm$ 0.194 & 14.643 & 11.197 & 14.555 & 22.625 \\ 
        ~ & 8 & 0.010 $\pm$ 0.000 & \textbf{4.200} $\pm$ 0.074 & \textbf{4.138} & 2.524 & \textbf{4.194} & 8.091 \\ 
        ~ & 9 & 0.010 $\pm$ 0.000 & 14.802 $\pm$ 0.199 & 14.678 & 10.361 & 14.700 & 23.429 \\ 
        ~ & 10 & 0.010 $\pm$ 0.000 & 4.419 $\pm$ 0.088 & 4.332 & 2.424 & 4.442 & 8.617 \\ \noalign{\vskip 1mm} \hline \noalign{\vskip 1mm}
        $10^{\times 5}$ & 0 & 0.025 $\pm$ 0.001 & 25.214 $\pm$ 0.275 & 25.063 & 18.646 & 25.012 & 33.507 \\ 
        ~ & 1 & 0.025 $\pm$ 0.001 & 5.512 $\pm$ 0.105 & 5.409 & 3.140 & 5.537 & 8.445 \\ 
        ~ & 2 & 0.025 $\pm$ 0.001 & 25.195 $\pm$ 0.276 & 25.041 & 17.925 & 25.227 & 33.771 \\ 
        ~ & 3 & 0.025 $\pm$ 0.001 & 25.598 $\pm$ 0.275 & 25.449 & 18.907 & 25.801 & 33.583 \\ 
        ~ & 4 & 0.025 $\pm$ 0.001 & 5.504 $\pm$ 0.113 & 5.383 & 3.106 & 5.585 & 8.536 \\ 
        ~ & 5 & 0.025 $\pm$ 0.001 & 6.391 $\pm$ 0.106 & 6.297 & 3.458 & 6.433 & 9.155 \\ 
        ~ & 6 & 0.025 $\pm$ 0.001 & 6.582 $\pm$ 0.109 & 6.487 & 3.697 & 6.590 & 9.444 \\ 
        ~ & 7 & 0.025 $\pm$ 0.001 & 24.640 $\pm$ 0.305 & 24.462 & 18.711 & 24.558 & 39.111 \\ 
        ~ & 8 & 0.025 $\pm$ 0.001 & \textbf{4.899} $\pm$ 0.068 & \textbf{4.850} & 3.126 & \textbf{4.914} & 6.599 \\ 
        ~ & 9 & 0.025 $\pm$ 0.001 & 24.502 $\pm$ 0.290 & 24.340 & 17.616 & 24.007 & 38.234 \\ 
        ~ & 10 & 0.025 $\pm$ 0.001 & 5.174 $\pm$ 0.084 & 5.106 & 3.203 & 5.127 & 8.583 \\ \noalign{\vskip 1mm} \hline \noalign{\vskip 1mm}
        $10^{\times 6}$ & 0 & 0.057 $\pm$ 0.002 & 33.437 $\pm$ 0.475 & 33.076 & 17.539 & 33.360 & 50.010 \\ 
        ~ & 1 & 0.057 $\pm$ 0.002 & 6.562 $\pm$ 0.111 & 6.467 & 3.647 & 6.566 & 9.963 \\ 
        ~ & 2 & 0.058 $\pm$ 0.002 & 33.606 $\pm$ 0.393 & 33.357 & 17.525 & 33.712 & 41.967 \\ 
        ~ & 3 & 0.058 $\pm$ 0.003 & 32.549 $\pm$ 0.436 & 32.241 & 18.479 & 32.502 & 43.975 \\ 
        ~ & 4 & 0.058 $\pm$ 0.002 & 6.883 $\pm$ 0.098 & 6.809 & 3.914 & 6.906 & 8.725 \\ 
        ~ & 5 & 0.058 $\pm$ 0.003 & 6.869 $\pm$ 0.112 & 6.770 & 4.069 & 7.110 & 9.002 \\ 
        ~ & 6 & 0.058 $\pm$ 0.003 & 7.310 $\pm$ 0.121 & 7.202 & 4.343 & 7.522 & 10.257 \\ 
        ~ & 7 & 0.057 $\pm$ 0.002 & 31.687 $\pm$ 0.419 & 31.394 & 16.671 & 31.747 & 42.899 \\ 
        ~ & 8 & 0.057 $\pm$ 0.002 & \textbf{5.750} $\pm$ 0.082 & \textbf{5.688} & 3.091 & \textbf{5.777} & 7.923 \\ 
        ~ & 9 & 0.057 $\pm$ 0.003 & 32.288 $\pm$ 0.399 & 32.024 & 18.471 & 32.798 & 41.716 \\ 
        ~ & 10 & 0.057 $\pm$ 0.003 & 6.313 $\pm$ 0.097 & 6.233 & 3.326 & 6.418 & 8.507 \\ \noalign{\vskip 1mm} \hline \noalign{\vskip 1mm}
        $10^{\times 7}$ & 0 & 0.145 $\pm$ 0.007 & 42.770 $\pm$ 1.192 & 41.208 & 15.333 & 41.701 & 96.280 \\ 
        ~ & 1 & 0.145 $\pm$ 0.007 & 8.799 $\pm$ 0.233 & \textbf{8.504} & 3.763 & \textbf{8.448} & 17.070 \\ 
        ~ & 2 & 0.147 $\pm$ 0.008 & 47.611 $\pm$ 1.045 & 46.352 & 17.773 & 46.859 & 78.931 \\ 
        ~ & 3 & 0.145 $\pm$ 0.008 & 42.546 $\pm$ 1.046 & 41.273 & 18.386 & 42.818 & 82.233 \\ 
        ~ & 4 & 0.147 $\pm$ 0.008 & 13.314 $\pm$ 0.253 & 13.054 & 6.425 & 13.584 & 19.319 \\ 
        ~ & 5 & 0.145 $\pm$ 0.008 & 9.650 $\pm$ 0.206 & 9.430 & 5.242 & 9.701 & 16.758 \\ 
        ~ & 6 & 0.145 $\pm$ 0.008 & 12.430 $\pm$ 0.254 & 12.168 & 6.576 & 12.144 & 20.844 \\ 
        ~ & 7 & 0.146 $\pm$ 0.008 & 41.994 $\pm$ 0.942 & 40.854 & 21.795 & 41.982 & 60.821 \\ 
        ~ & 8 & 0.146 $\pm$ 0.008 & \textbf{8.729} $\pm$ 0.190 & 8.522 & 5.101 & 8.807 & 14.279 \\ 
        ~ & 9 & 0.146 $\pm$ 0.008 & 46.652 $\pm$ 1.067 & 45.448 & 21.751 & 46.404 & 83.582 \\ 
        ~ & 10 & 0.146 $\pm$ 0.008 & 11.801 $\pm$ 0.254 & 11.534 & 6.208 & 11.542 & 20.518 \\ \noalign{\vskip 1mm} \hline \noalign{\vskip 1mm}
        $10^{\times 8}$ & 0 & 0.381 $\pm$ 0.021 & 96.221 $\pm$ 5.042 & 83.698 & 15.480 & 88.210 & 333.633 \\ 
        ~ & 1 & 0.381 $\pm$ 0.021 & \textbf{42.482} $\pm$ 2.792 & 36.519 & 6.272 & 38.715 & 254.793 \\ 
        ~ & 2 & 0.387 $\pm$ 0.021 & 140.427 $\pm$ 10.279 & 121.552 & 22.769 & 125.905 & 820.344 \\ 
        ~ & 3 & 0.381 $\pm$ 0.021 & 115.342 $\pm$ 18.396 & 82.210 & 14.612 & 84.227 & 1690.488 \\ 
        ~ & 4 & 0.387 $\pm$ 0.021 & 80.225 $\pm$ 7.295 & 68.880 & 13.509 & 68.795 & 588.859 \\ 
        ~ & 5 & 0.381 $\pm$ 0.021 & 62.368 $\pm$ 17.093 & \textbf{34.881} & 6.601 & \textbf{33.880} & 1589.041 \\ 
        ~ & 6 & 0.380 $\pm$ 0.021 & 82.748 $\pm$ 14.217 & 56.711 & 10.151 & 54.676 & 1121.405 \\ 
        ~ & 7 & 0.386 $\pm$ 0.021 & 106.177 $\pm$ 8.930 & 86.913 & 15.541 & 85.482 & 678.892 \\ 
        ~ & 8 & 0.386 $\pm$ 0.021 & 51.834 $\pm$ 7.228 & 38.044 & 6.814 & 37.735 & 519.459 \\ 
        ~ & 9 & 0.385 $\pm$ 0.021 & 124.828 $\pm$ 9.714 & 105.228 & 19.539 & 111.054 & 923.472 \\ 
        ~ & 10 & 0.385 $\pm$ 0.021 & 70.414 $\pm$ 8.270 & 57.581 & 10.774 & 58.946 & 837.845 \\ \noalign{\vskip 1mm} \hline \noalign{\vskip 1mm}    
        \multicolumn{8}{p{15cm}}{Note: For each problem, the lowest mean and median computation time are highlighted.}  
\end{longtable}
\end{center}

\newpage

\begin{center}
\begin{longtable}{c|c|c|ccccc}
    \caption{Results for nonnegative tensors with size $10^{\times 6}$ and increasing $n$ samples}\label{inc-n-1} \\
    \noalign{\vskip 1mm} \hline \noalign{\vskip 1mm}     
    \textbf{Sample} & \textbf{Version} & \textbf{NMSE} & \multicolumn{5}{c}{\textbf{Time (s)}} \\ 
    \textbf{Percent} & ~ & Arithmetic & Arithmetic & Geometric  &  &  & \\
    ~ & ~ & Mean $\pm$ SE & Mean $\pm$ SE &  Mean & Min & Median & Max \\ \noalign{\vskip 1mm} \hline \noalign{\vskip 1mm}    
    0.01 & 0 & 0.982 $\pm$ 0.004 & 0.276 $\pm$ 0.010 & 0.258 & 0.142 & 0.246 & 0.650 \\ 
        ~ & 1 & 0.982 $\pm$ 0.004 & \textbf{0.243} $\pm$ 0.009 & \textbf{0.227} & 0.122 & 0.215 & 0.546 \\ 
        ~ & 2 & 0.981 $\pm$ 0.004 & 0.329 $\pm$ 0.016 & 0.298 & 0.150 & 0.285 & 1.019 \\ 
        ~ & 3 & 0.982 $\pm$ 0.004 & 0.304 $\pm$ 0.012 & 0.282 & 0.148 & 0.265 & 0.649 \\ 
        ~ & 4 & 0.981 $\pm$ 0.004 & 0.299 $\pm$ 0.015 & 0.270 & 0.130 & 0.256 & 0.919 \\ 
        ~ & 5 & 0.982 $\pm$ 0.004 & 0.265 $\pm$ 0.011 & 0.246 & 0.129 & 0.242 & 0.576 \\ 
        ~ & 6 & 0.982 $\pm$ 0.004 & 0.292 $\pm$ 0.013 & 0.266 & 0.131 & 0.281 & 0.685 \\ 
        ~ & 7 & 0.980 $\pm$ 0.004 & 0.284 $\pm$ 0.012 & 0.264 & 0.151 & 0.252 & 0.654 \\ 
        ~ & 8 & 0.980 $\pm$ 0.004 & 0.250 $\pm$ 0.010 & 0.232 & 0.131 & \textbf{0.211} & 0.549 \\ 
        ~ & 9 & 0.981 $\pm$ 0.004 & 0.328 $\pm$ 0.016 & 0.299 & 0.148 & 0.281 & 0.873 \\ 
        ~ & 10 & 0.981 $\pm$ 0.004 & 0.293 $\pm$ 0.015 & 0.264 & 0.129 & 0.253 & 0.817 \\ \noalign{\vskip 1mm} \hline \noalign{\vskip 1mm}    
        0.1 & 0 & 0.564 $\pm$ 0.022 & 9.507 $\pm$ 0.439 & 8.565 & 2.548 & 8.182 & 21.995 \\ 
        ~ & 1 & 0.564 $\pm$ 0.022 & 2.977 $\pm$ 0.116 & 2.758 & 0.843 & \textbf{2.616} & 6.217 \\ 
        ~ & 2 & 0.568 $\pm$ 0.022 & 9.161 $\pm$ 0.370 & 8.469 & 2.804 & 8.037 & 20.648 \\ 
        ~ & 3 & 0.571 $\pm$ 0.022 & 9.504 $\pm$ 0.415 & 8.647 & 2.548 & 8.306 & 24.41 \\ 
        ~ & 4 & 0.568 $\pm$ 0.022 & 3.176 $\pm$ 0.107 & 3.002 & 1.087 & 2.948 & 6.223 \\ 
        ~ & 5 & 0.571 $\pm$ 0.022 & 3.040 $\pm$ 0.116 & 2.822 & 0.874 & 2.836 & 7.308 \\ 
        ~ & 6 & 0.571 $\pm$ 0.022 & 3.482 $\pm$ 0.130 & 3.237 & 0.952 & 3.273 & 8.053 \\ 
        ~ & 7 & 0.547 $\pm$ 0.023 & 9.270 $\pm$ 0.467 & 8.291 & 2.227 & 8.018 & 31.432 \\ 
        ~ & 8 & 0.547 $\pm$ 0.023 & \textbf{2.931} $\pm$ 0.140 & \textbf{2.666} & 0.844 & 2.669 & 10.910 \\ 
        ~ & 9 & 0.562 $\pm$ 0.022 & 9.111 $\pm$ 0.388 & 8.329 & 2.841 & 8.463 & 21.631 \\ 
        ~ & 10 & 0.562 $\pm$ 0.022 & 3.334 $\pm$ 0.129 & 3.089 & 0.965 & 3.194 & 7.059 \\ \noalign{\vskip 1mm} \hline \noalign{\vskip 1mm}    
        1 & 0 & 0.057 $\pm$ 0.002 & 33.437 $\pm$ 0.475 & 33.076 & 17.539 & 33.360 & 50.010 \\ 
        ~ & 1 & 0.057 $\pm$ 0.002 & 6.562 $\pm$ 0.111 & 6.467 & 3.647 & 6.566 & 9.963 \\ 
        ~ & 2 & 0.058 $\pm$ 0.002 & 33.606 $\pm$ 0.393 & 33.357 & 17.525 & 33.712 & 41.967 \\ 
        ~ & 3 & 0.058 $\pm$ 0.003 & 32.549 $\pm$ 0.436 & 32.241 & 18.479 & 32.502 & 43.975 \\ 
        ~ & 4 & 0.058 $\pm$ 0.002 & 6.883 $\pm$ 0.098 & 6.809 & 3.914 & 6.906 & 8.725 \\ 
        ~ & 5 & 0.058 $\pm$ 0.003 & 6.869 $\pm$ 0.112 & 6.770 & 4.069 & 7.110 & 9.002 \\ 
        ~ & 6 & 0.058 $\pm$ 0.003 & 7.310 $\pm$ 0.121 & 7.202 & 4.343 & 7.522 & 10.257 \\ 
        ~ & 7 & 0.057 $\pm$ 0.002 & 31.687 $\pm$ 0.419 & 31.394 & 16.671 & 31.747 & 42.899 \\ 
        ~ & 8 & 0.057 $\pm$ 0.002 & \textbf{5.750} $\pm$ 0.082 & \textbf{5.688} & 3.091 & \textbf{5.777} & 7.923 \\ 
        ~ & 9 & 0.057 $\pm$ 0.003 & 32.288 $\pm$ 0.399 & 32.024 & 18.471 & 32.798 & 41.716 \\ 
        ~ & 10 & 0.057 $\pm$ 0.003 & 6.313 $\pm$ 0.097 & 6.233 & 3.326 & 6.418 & 8.507 \\ \noalign{\vskip 1mm} \hline \noalign{\vskip 1mm}    
        10 & 0 & 0.050 $\pm$ 0.002 & 266.905 $\pm$ 3.785 & 263.971 & 138.684 & 266.407 & 370.278 \\ 
        ($n=10^5$) & 1 & 0.050 $\pm$ 0.002 & 58.075 $\pm$ 1.347 & 56.526 & 31.177 & 56.269 & 101.203 \\ 
        ~ & 2 & 0.051 $\pm$ 0.002 & 269.866 $\pm$ 3.637 & 267.083 & 133.925 & 268.477 & 359.523 \\ 
        ~ & 3 & 0.050 $\pm$ 0.002 & 246.770 $\pm$ 3.353 & 244.199 & 138.894 & 249.96 & 335.414 \\ 
        ~ & 4 & 0.051 $\pm$ 0.002 & 58.629 $\pm$ 1.302 & 57.232 & 33.940 & 56.590 & 93.109 \\ 
        ~ & 5 & 0.050 $\pm$ 0.002 & 56.653 $\pm$ 0.964 & 55.822 & 37.055 & 56.013 & 80.008 \\ 
        ~ & 6 & 0.050 $\pm$ 0.002 & 56.285 $\pm$ 0.975 & 55.431 & 34.631 & 55.497 & 79.487 \\ 
        ~ & 7 & 0.050 $\pm$ 0.002 & 241.195 $\pm$ 3.388 & 238.465 & 138.051 & 242.977 & 333.482 \\ 
        ~ & 8 & 0.050 $\pm$ 0.002 & \textbf{48.434} $\pm$ 0.696 & \textbf{47.934} & 30.473 & \textbf{48.433} & 69.063 \\ 
        ~ & 9 & 0.050 $\pm$ 0.002 & 244.385 $\pm$ 3.051 & 242.218 & 149.075 & 245.842 & 335.246 \\ 
        ~ & 10 & 0.050 $\pm$ 0.002 & 50.598 $\pm$ 0.711 & 50.099 & 34.797 & 50.104 & 73.951 \\ \noalign{\vskip 1mm} \hline \noalign{\vskip 1mm}    
        \multicolumn{8}{p{15cm}}{Note: For each problem, the lowest mean and median computation time are highlighted.}  
\end{longtable}
\end{center}

\newpage

\begin{center}
\begin{longtable}{c|c|c|ccccc}
    \caption{Results for nonnegative tensors with size $10^{\times 7}$ and increasing $n$ samples}\label{inc-n-2} \\
    \noalign{\vskip 1mm} \hline \noalign{\vskip 1mm}     
    \textbf{Sample} & \textbf{Version} & \textbf{NMSE} & \multicolumn{5}{c}{\textbf{Time (s)}} \\ 
    \textbf{Percent} & ~ & Arithmetic & Arithmetic & Geometric  &  &  & \\
    ~ & ~ & Mean $\pm$ SE & Mean $\pm$ SE &  Mean & Min & Median & Max \\ \noalign{\vskip 1mm} \hline \noalign{\vskip 1mm}    
        0.01 & 0 & 0.884 $\pm$ 0.016 & 9.906 $\pm$ 0.733 & 7.952 & 1.338 & 7.456 & 44.304 \\ 
        ~ & 1 & 0.884 $\pm$ 0.016 & 5.154 $\pm$ 0.371 & 4.266 & 0.772 & 3.993 & 25.657 \\ 
        ~ & 2 & 0.883 $\pm$ 0.016 & 10.435 $\pm$ 0.717 & 8.677 & 1.413 & 8.409 & 41.956 \\ 
        ~ & 3 & 0.885 $\pm$ 0.015 & 8.978 $\pm$ 0.637 & 7.337 & 1.441 & 7.069 & 35.707 \\ 
        ~ & 4 & 0.883 $\pm$ 0.016 & 6.120 $\pm$ 0.385 & 5.259 & 0.897 & 5.237 & 25.234 \\ 
        ~ & 5 & 0.885 $\pm$ 0.015 & 4.542 $\pm$ 0.307 & 3.846 & 0.833 & 3.825 & 20.304 \\ 
        ~ & 6 & 0.888 $\pm$ 0.015 & 6.185 $\pm$ 0.496 & 5.055 & 0.913 & 4.931 & 34.564 \\ 
        ~ & 7 & 0.870 $\pm$ 0.018 & 8.363 $\pm$ 0.612 & 6.796 & 1.506 & 6.664 & 40.259 \\ 
        ~ & 8 & 0.870 $\pm$ 0.018 & \textbf{4.242} $\pm$ 0.286 & \textbf{3.585} & 0.818 & \textbf{3.483} & 18.221 \\ 
        ~ & 9 & 0.874 $\pm$ 0.017 & 9.320 $\pm$ 0.663 & 7.599 & 1.565 & 7.122 & 32.849 \\ 
        ~ & 10 & 0.874 $\pm$ 0.017 & 5.444 $\pm$ 0.373 & 4.546 & 0.939 & 4.325 & 21.065 \\ \noalign{\vskip 1mm} \hline \noalign{\vskip 1mm}  
        0.1 & 0 & 0.145 $\pm$ 0.007 & 42.770 $\pm$ 1.192 & 41.208 & 15.333 & 41.701 & 96.280 \\ 
        ~ & 1 & 0.145 $\pm$ 0.007 & 8.799 $\pm$ 0.233 & \textbf{8.504} & 3.763 & \textbf{8.448} & 17.070 \\ 
        ~ & 2 & 0.147 $\pm$ 0.008 & 47.611 $\pm$ 1.045 & 46.352 & 17.773 & 46.859 & 78.931 \\ 
        ~ & 3 & 0.145 $\pm$ 0.008 & 42.546 $\pm$ 1.046 & 41.273 & 18.386 & 42.818 & 82.233 \\ 
        ~ & 4 & 0.147 $\pm$ 0.008 & 13.314 $\pm$ 0.253 & 13.054 & 6.425 & 13.584 & 19.319 \\ 
        ~ & 5 & 0.145 $\pm$ 0.008 & 9.650 $\pm$ 0.206 & 9.430 & 5.242 & 9.701 & 16.758 \\ 
        ~ & 6 & 0.145 $\pm$ 0.008 & 12.430 $\pm$ 0.254 & 12.168 & 6.576 & 12.144 & 20.844 \\ 
        ~ & 7 & 0.146 $\pm$ 0.008 & 41.994 $\pm$ 0.942 & 40.854 & 21.795 & 41.982 & 60.821 \\ 
        ~ & 8 & 0.146 $\pm$ 0.008 & \textbf{8.729} $\pm$ 0.190 & 8.522 & 5.101 & 8.807 & 14.279 \\ 
        ~ & 9 & 0.146 $\pm$ 0.008 & 46.652 $\pm$ 1.067 & 45.448 & 21.751 & 46.404 & 83.582 \\ 
        ~ & 10 & 0.146 $\pm$ 0.008 & 11.801 $\pm$ 0.254 & 11.534 & 6.208 & 11.542 & 20.518 \\ \noalign{\vskip 1mm} \hline \noalign{\vskip 1mm}  
        1 & 0 & 0.112 $\pm$ 0.005 & 245.863 $\pm$ 6.133 & 238.338 & 148.868 & 233.195 & 408.085 \\ 
        ~ & 1 & 0.112 $\pm$ 0.005 & 47.923 $\pm$ 0.899 & 47.145 & 32.281 & \textbf{45.655} & 77.549 \\ 
        ~ & 2 & 0.112 $\pm$ 0.005 & 252.129 $\pm$ 5.768 & 245.421 & 148.470 & 248.945 & 366.094 \\ 
        ~ & 3 & 0.113 $\pm$ 0.005 & 247.820 $\pm$ 5.260 & 242.007 & 145.381 & 264.145 & 350.913 \\ 
        ~ & 4 & 0.112 $\pm$ 0.005 & 53.755 $\pm$ 0.978 & 52.931 & 35.569 & 52.759 & 95.705 \\ 
        ~ & 5 & 0.113 $\pm$ 0.005 & 53.408 $\pm$ 0.811 & 52.784 & 33.295 & 53.193 & 73.774 \\ 
        ~ & 6 & 0.113 $\pm$ 0.005 & 55.586 $\pm$ 0.824 & 54.963 & 34.203 & 55.012 & 76.677 \\ 
        ~ & 7 & 0.110 $\pm$ 0.005 & 233.996 $\pm$ 5.571 & 227.430 & 143.104 & 233.493 & 388.936 \\ 
        ~ & 8 & 0.110 $\pm$ 0.005 & \textbf{46.640} $\pm$ 0.647 & \textbf{46.189} & 32.440 & 46.543 & 64.163 \\ 
        ~ & 9 & 0.109 $\pm$ 0.005 & 238.510 $\pm$ 5.316 & 232.407 & 144.561 & 235.277 & 326.239 \\ 
        ~ & 10 & 0.109 $\pm$ 0.005 & 50.178 $\pm$ 0.679 & 49.708 & 34.546 & 50.580 & 65.287 \\ \noalign{\vskip 1mm} \hline \noalign{\vskip 1mm}  
        10 & 0 & 0.108 $\pm$ 0.005 & 2830.535 $\pm$ 57.672 & 2773.083 & 1868.109 & 2654.533 & 4015.562 \\ 
        ($n=10^6$) & 1 & 0.108 $\pm$ 0.005 & 1223.087 $\pm$ 13.412 & 1216.033 & 989.056 & 1225.745 & 1662.859 \\ 
        ~ & 2 & 0.108 $\pm$ 0.005 & 2776.876 $\pm$ 53.723 & 2726.519 & 1940.371 & 2576.609 & 3907.286 \\ 
        ~ & 3 & 0.110 $\pm$ 0.005 & 2781.695 $\pm$ 55.039 & 2728.281 & 1899.844 & 2749.232 & 4377.475 \\ 
        ~ & 4 & 0.108 $\pm$ 0.005 & 1138.666 $\pm$ 27.790 & 1117.027 & 910.320 & 1065.029 & 2873.461 \\ 
        ~ & 5 & 0.110 $\pm$ 0.005 & 1182.486 $\pm$ 19.024 & 1170.131 & 939.523 & 1139.444 & 2230.349 \\ 
        ~ & 6 & 0.110 $\pm$ 0.005 & 1137.827 $\pm$ 10.332 & 1133.184 & 917.410 & 1116.388 & 1408.491 \\ 
        ~ & 7 & 0.108 $\pm$ 0.005 & 2636.415 $\pm$ 54.039 & 2583.335 & 1730.813 & 2380.683 & 3738.891 \\ 
        ~ & 8 & 0.108 $\pm$ 0.005 & \textbf{1045.224} $\pm$ 8.729 & \textbf{1041.659} & 899.932 & \textbf{1032.678} & 1261.093 \\ 
        ~ & 9 & 0.110 $\pm$ 0.005 & 2687.159 $\pm$ 50.578 & 2640.935 & 1957.054 & 2510.885 & 3761.409 \\ 
        ~ & 10 & 0.110 $\pm$ 0.005 & 1090.345 $\pm$ 8.418 & 1087.132 & 928.105 & 1076.160 & 1310.540 \\ \noalign{\vskip 1mm} \hline \noalign{\vskip 1mm}     
        \multicolumn{8}{p{15cm}}{Note: For each problem, the lowest mean and median computation time are highlighted.}
\end{longtable}
\end{center}

\end{document}